\documentclass{article}
\pdfoutput=1
\usepackage{graphicx}
\usepackage{algorithm}
\usepackage{algorithmic}
\usepackage{newfloat}
\usepackage{listings}
\usepackage{amssymb}
\usepackage{diagbox}
\graphicspath{{./}}
\usepackage{doi}

\newcommand{\HRule}[1]{\rule{\linewidth}{#1}}

\begin{document}
	
\title{
	\HRule{2pt} \\
	\textbf{MDM: Multiple Dynamic Masks for Visual Explanation of Neural Networks}
	\HRule{0.5pt}
	}

\author{
	\textbf{Yitao Peng},
	\textbf{Longzhen Yang},
	\textbf{Yihang Liu},
	\textbf{Lianghua He\textsuperscript{*}}\\
	College of Electronic and Information Engineering Tongji University\\
	4800 Cao’an Highway, Shanghai, China 201804\\
	\{pyt, yanglongzhen, 2111131, helianghua\}@tongji.edu.cn
}
\date{}

\maketitle	

\begin{abstract}
The Class Activation Map (CAM) lookup of a neural network tells us to which regions the neural network focuses when it makes a decision. In the past, the CAM search method was dependent upon a specific internal module of the network. It has specific constraints on the structure of the neural network. To make the search of CAM have generality and high performance. We propose a learning-based algorithm, namely Multiple Dynamic Masks (MDM). It is based on a public cognition that only active features of a picture related to classification will affect the classification results of the neural network, and other features will hardly affect the classification results of the network. The mask generated by MDM conforms to the above cognition. It trains mask vectors of different sizes by constraining mask values and activating consistency, then it uses stacking masks of different scale to generate CAM that can balance spatial information and semantic information. Comparing the results of MDM with those of the recent advanced CAM search method, the performance of MDM has reached the state of the art results. We applied the MDM method to the interpretable neural networks ProtoPNet and XProtoNet, which improved the performance of model in the explainable prototype search. Finally, we visualized the CAM generation effect of MDM on neural networks of different architectures, verifying the generality of the MDM method.
\end{abstract}

\thispagestyle{empty}
\newpage
\setcounter{page}{2}

\section{Introduction}

Neural networks \cite{simonyan2014very,szegedy2015going,he2016deep,huang2017densely,dosovitskiy2020image, liu2021swin,liu2022convnet} have achieved remarkable success in the field of image classification, and the explanation of neural network has started to gain attention.

Since the neural network is a black-box model, humans cannot understand the decision-making of the neural network. Hence, it is difficult to build trust on such models, thus making it difficult to apply the neural network to some important fields, such as the pathological diagnosis, and unmanned driving. In the medical field, it is a major issue whether the diagnosis results of neural networks related to life are credible or not. Although many models \cite{hashimoto2020multi,maksoud2020sos} can achieve good results in the field of medical diagnosis and even surpass human doctors, neither doctors nor patients can trust the models without a good explanation of the model's decisions \cite{patricio2022explainable}. This is the reason why neural networks even with better classification performance still cannot be used in the clinical field.

Based on the above requirements, various methods \cite{zhou2016learning,selvaraju2017grad,shrikumar2017learning,sundararajan2017axiomatic,chattopadhay2018grad,petsiuk2018rise,wang2020score,ramaswamy2020ablation} have been proposed for explaining neural network models. According to \cite{patricio2022explainable}, the methods of explaining neural network models can be divided into two categories: explaining the reasoning process of the neural network and explaining the information of the neural network. The explanation process is to make the inference process of the model conform to human cognition. The interpretation of the information of a neural network module involves the study of the gradients of its some modules and the activation of internal hidden layers to determine which areas of the image are promoting the neural network to make decisions.

\begin{figure}[t]
	\centering
	\includegraphics[width=0.9\columnwidth]{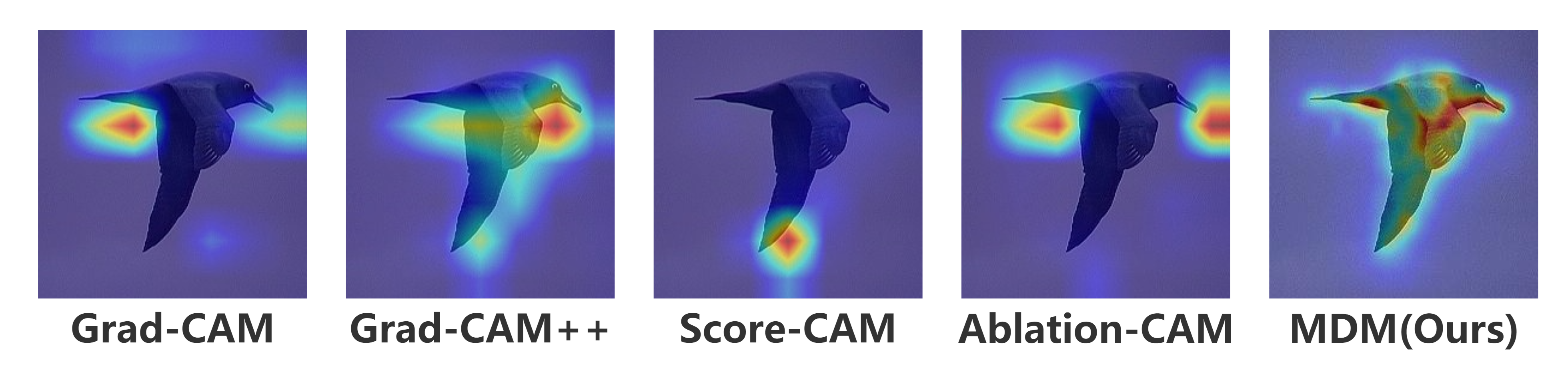} 
	\caption{Class Activation Map generate from different models. From blue to red, the activation degree increase.}
	\label{fig1}
\end{figure}

The above neural network interpretation methods have their own limitations. There are \cite{selvaraju2017grad,zhou2016learning,chattopadhay2018grad,wang2020score,ramaswamy2020ablation} methods for interpreting the information of a neural network module, which require specific restrictions on the internal structure of the neural network. For example, CAM needs to capture the global average of the pooling layer information at the end of the neural network. This approach is difficult to apply to recent state-of-the-art neural network models, such as those of the Transformer architecture \cite{dosovitskiy2020image,liu2021swin}. These methods are not universal.

The neural network ProtoPNet \cite{chen2019looks}, XProtoNet \cite{kim2021xprotonet} and other models for classification of the interpretability of the reasoning process need to be set to a specific interpretable architecture, which degrade the network performance.

The neural network makes the correct classification based on the foreground information, which is essential and invariant. The background information belongs to auxiliary information or redundant information that can be changed arbitrarily. This is the fact of human cognition.

In order to fit this assumption, we set up multiple mask feature vectors, each vector is upsampled to generate a mask, and the corresponding activation values of the original image and the masked image are calculated on the neural network. The structure of MDM is shown in Figure 2. MDM preserves the picture information that is helpful for neural network decision-making and removes redundant information by maximizing the consistency of this activation while keeping the mask value as small as possible through a learning-based method. MDM sets mask feature vectors of different sizes, and each element of the mask feature vectors of different sizes corresponds to receptive fields of different sizes. The feature vector corresponding to a larger receptive field can retain better semantic information, so that the mask value corresponding to each region is accurate. While the ability of the mask feature vector to express spatial information is poor, the fine-grained division of the image is imprecise. Conversely, feature vectors have poor semantic information, but can better preserve spatial information. Therefore, MDM allows the CAM to preserve the spatial information and semantic information of the original image by stacking masks of different sizes, which also makes the CAM robust.

Since MDM regards the neural network structure as a black box and it does not use the structural information inside the network, it can be applied to any structure of the neural network with good generality.

MDM is a learning-based method, which has adaptability to input data and network, thus it has excellent performance for explaining the classification basis of neural network. The results are shown in Figure 1, in our multiple experiments comparing the most advanced models of recent years, MDM achieves the state-of-the-art in all the indicators.

The main contributions of this paper can be summarized as follows:

\begin{itemize}
    \item We propose a general, interpretable and excellent method for finding basis for classification decisions in neural networks. We give a mathematical proof of the feasibility of MDM algorithm under certain conditions.
    \item We implement the MDM through mask fusion of multi-scale receptive fields and learning-based methods. We verified that the performance of the MDM method achieved the state of the art result in the search of decision area of neural networks for classification.
    \item We apply the MDM method to the interpretable neural networks ProtoPNet and XProtoNet of traditional image datasets and  pathological image datasets, which greatly improved the search prototype performance of the above interpretable neural networks in classifying traditional images and medical images.
    \item We test the MDM method to the current advanced convolutional neural networks ResNet, VGG, DenseNet, VIT and Swin-Transformer. The decision basis for the classification of the neural network is visualized. We validate the general applicability of the MDM method to these advanced and structurally different neural networks.
\end{itemize}

\section{Related Work}

\subsection{Saliency Maps}
CAM allowed the generation of saliency maps. In the Global Average Pooling (GAP) proposed by Lin \cite{lin2013network}, GAP integrates the information of all the features of the whole space. Zhou \cite{zhou2016learning} proposed to use GAP for obtaining the CAM. Selvaraju \cite{selvaraju2017grad} proposed Grad-CAM, which needs to use the gradient only. Obtaining activation maps makes the method more universal in obtaining CAM. Shrikumar \cite{shrikumar2017learning} proposed Deep Learning Important FeaTures (DeepLIFT) to decompose the neural network's output prediction. Sundararajan \cite{sundararajan2017axiomatic} proposed Integrated Gradients (IG) to find the input features for the prediction attribution of deep networks. Chattopadhay \cite{chattopadhay2018grad} proposed Grad-CAM++, which  added an additional weight to weigh the elements of the gradient map. It makes the CAM localization more accurate. In 2020, Wang \cite{wang2020score} proposed a gradient-free approach, namely Score-CAM. To a certain extent, the problems of the neural network gradient noise, saturation and easy to find false confidence samples are solved. Ramaswamy \cite{ramaswamy2020ablation} proposed Ablation-CAM, which explore the strength of each factor's contribution to the overall model, and find the most important factor affecting performance. The derivation of above methods are not easy for humans to understand, and it is not convenient to migrate from one model to another new model. For example, the above methods are not easy to transfer to the network of Transformer architecture. Therefore, the above methods lack generality.

\subsection{Interpretable Models}

Interpretability based on inference process. Petsiuk \cite{petsiuk2018rise} proposed a neural network interpretability analysis method, namely RISE. The fitting space of the model is small, thus its representation ability is insufficient. Chen \cite{chen2019looks} proposed ProtoPNet, which infers the category of the input data by finding whether the input picture has a similar prototype for a certain category. Kim \cite{kim2021xprotonet} proposed an improved network XProtoNet to set the prototype as a feature vector with variable activation location and size, and applied it to the interpretable classification of X-ray images of chest diseases. Singh \cite{singh2021interpretable,singh2021these} proposed networks NP-ProtoPNet and Gen-ProtoPNet based on ProtoPNet, which further generalized the size of feature vectors representing prototypes.

The network with the above ProtoPNet structure upsamples the prototype activation feature map into an activation map of the original image size, and uses its larger activation area as the neural network classification basis. Direct upsampling  method of extracting prototypes is not interpretable and this method has a moderate lookup performance. The MDM algorithm proposed in this paper can be applied to these networks. As a network prototype search method, it not only greatly improves the prototype search performance, but also the search process is interpretable.

\begin{figure*}[t]
	\centering
	\includegraphics[width=0.8\textwidth]{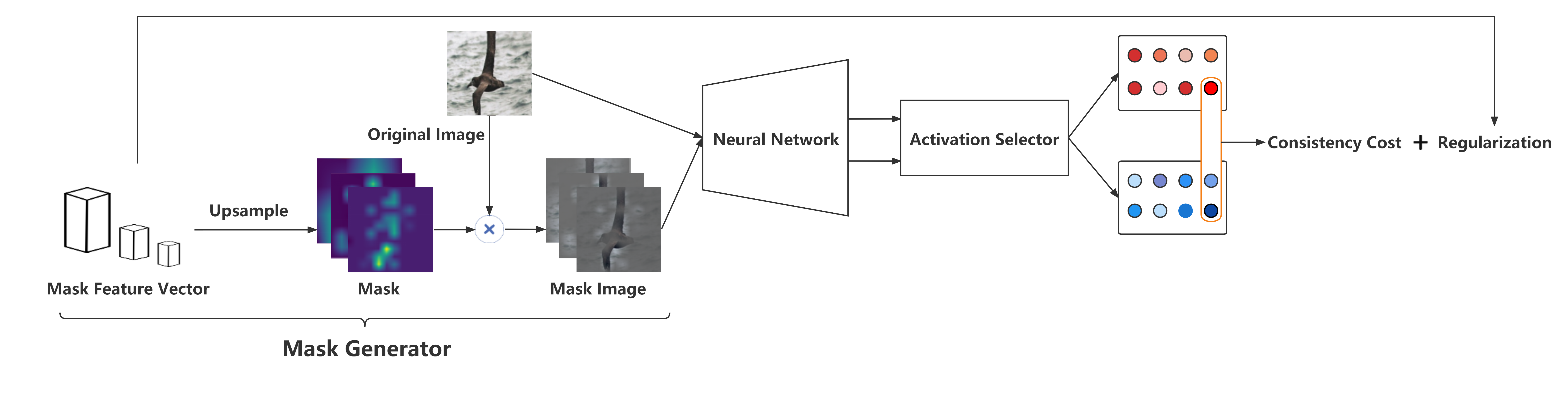} 
	\caption{Overall architecture of Multiple Dynamic Masks to train mask vectors.}
	\label{fig2}
\end{figure*}

\section{Method}

\subsection{Architecture and Objective Function}

MDM consists of mask generator, neural network and activation selector. It minimizes the value of masks while preserves the decision information in favor of classification. The objective functions for the mask training and mask generating are represented as follows. The meanings of letters and symbols are explained by the following subsections.

Train function:
\begin{equation}
\min\limits_{d_{i}}{||f_{p}(X)-f_{p}(g(d_{i})X)||^2}+\frac{\lambda_{i}}{|a_{i}b_{i}|}\sum_{u=1}^{a_i}{\sum_{v=1}^{b_i}|d_{iuv}|}
\end{equation}

Generation function:

\begin{equation}
M^{h}=Normalize(\{\sum_{i=1}^{N}g(d_{i})\geq\gamma\}\sum_{i=1}^{N}g(d_{i}))
\end{equation}

\subsubsection{Mask Generator.}
Set the appropriate $N$ mask feature vectors $D\{d_{i}\}^{N}_{i=1}$ to be trained for the dataset, $d_{i}\in R^{a_{i}\times b_{i} \times 1}$, where each value in $d_{i}$ is set as a fixed value. For any $i,j\in\{1,2,..,N\}$, if $i \neq j$ then $a_{i} \neq a_{j}$ or $b_{i} \neq b_{j}$ . Select the mask feature vector transformation function $g(\cdot)$ to generate mask $M_{i}=g(d_{i})\in R^{H\times W\times 1}$, where $g(\cdot)$ contains the normalized operation, so that each element value in $M_{i}$ belongs to $[0, 1]$. $M_{i}X$ and $X$ are inputs for calculating the activation consistency of neural network $f$. $d_{i}$ and mask $M_{i}=g(d_{i})$ are obtained by minimizing the objective function (1).

As shown in Figure 3, the smaller the size of $d_{i}$, the larger is the receptive field corresponding to each element point in $d_{i}$ to contain more semantic information, and the more accurate $f$ makes decisions. (1) The performance of the training mask is better, but the disadvantages are that the granularity of the activation area division in the mask is too large, the spatial information is less, and the spatial feature division is not fine. On the contrary, when the size of $d_{i}$ is larger, the corresponding receptive field is smaller, and the mask division space area is more refined, but the semantic information of each receptive field is less, the activation accuracy of the calculated activation area is lower, and the mask performance is relatively normal.

We need to consider the accuracy of both semantic information and spatial information. Therefore, upon stacking $d_{i}$ masks of various sizes, the generated masks can balance the above two accuracies and obtain robustness.

When the receptive field corresponding to a single element point of $d_{i}$ is too small, the classification performance of the neural network $f$ is not good, and an adversarial effect occurs \cite{petsiuk2018rise}. Therefore, we set a lower threshold to remove the activation areas smaller than the threshold, and remove these redundant or erroneous ones. The mask calculation formula is given in (2).

\begin{figure}[t]
	\centering
	\includegraphics[width=0.9\columnwidth]{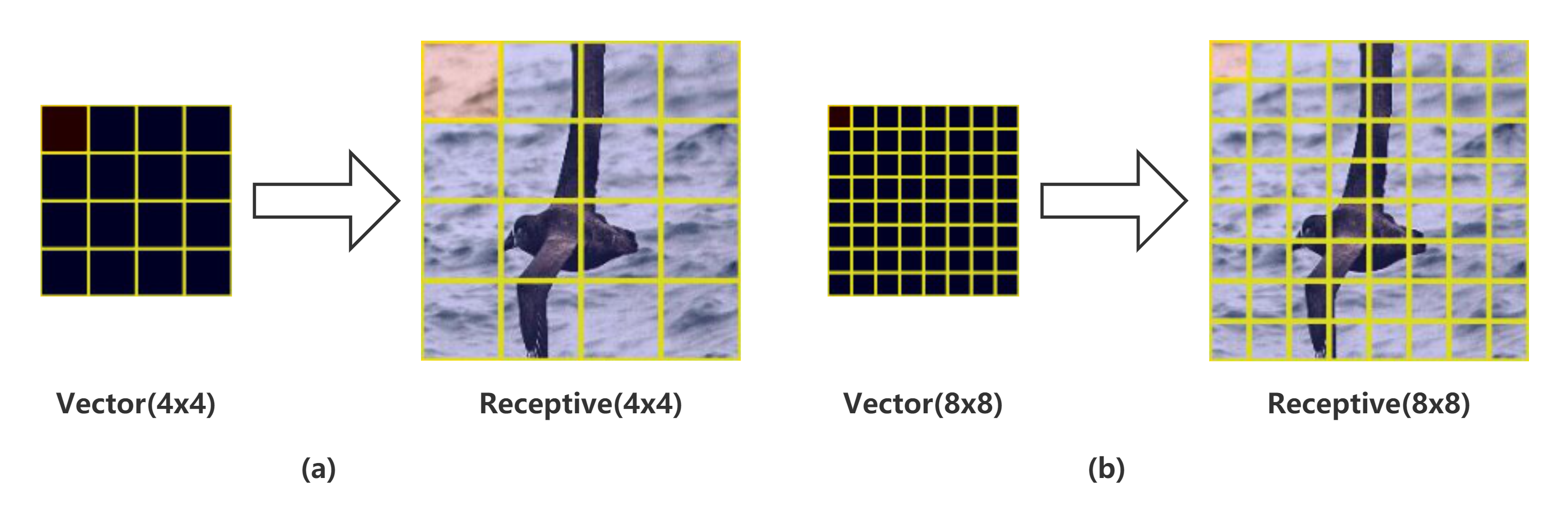} 
	\caption{After $d_{i}$ of different sizes are upsampled to the original image size, the red area represents the receptive field corresponding to each pixel point in $d_{i}$. (a) indicates that  $d_{i}$ is a 4×4 size mask feature vector, and each pixel in $d_{i}$ corresponds to a 56×56 size area (red area) in the original image after upsampling. In (b), $d_{i}$ is a mask feature vector of size 8×8, and each pixel in $d_{i}$ corresponds to a 28×28 area (red area) in the original image after upsampling.}
	\label{fig3}
\end{figure}

\subsubsection{Neural Network.}
Just put the trained neural network into this module, keep the parameters of the neural network unchanged, and use the output of the mask generator as the input. The activation selector selects some nodes of the neural network as activation regions.

\subsubsection{Activation Selector.}
The activation selector module selects the specific activation position of the neural network, so that the original image and the mask image are activated consistently in the activation position to train the mask vector. For example, in classfication task, $X$ is the input image, $f$ is neural network. $f(X) \triangleq t$, activation position is $t$. For ProtoPNet \cite{chen2019looks} and XProtoNet \cite{kim2021xprotonet}, their location of prototype tensor is chosen as activation position.

\begin{figure*}[t]
	\centering
	\includegraphics[width=0.8\textwidth]{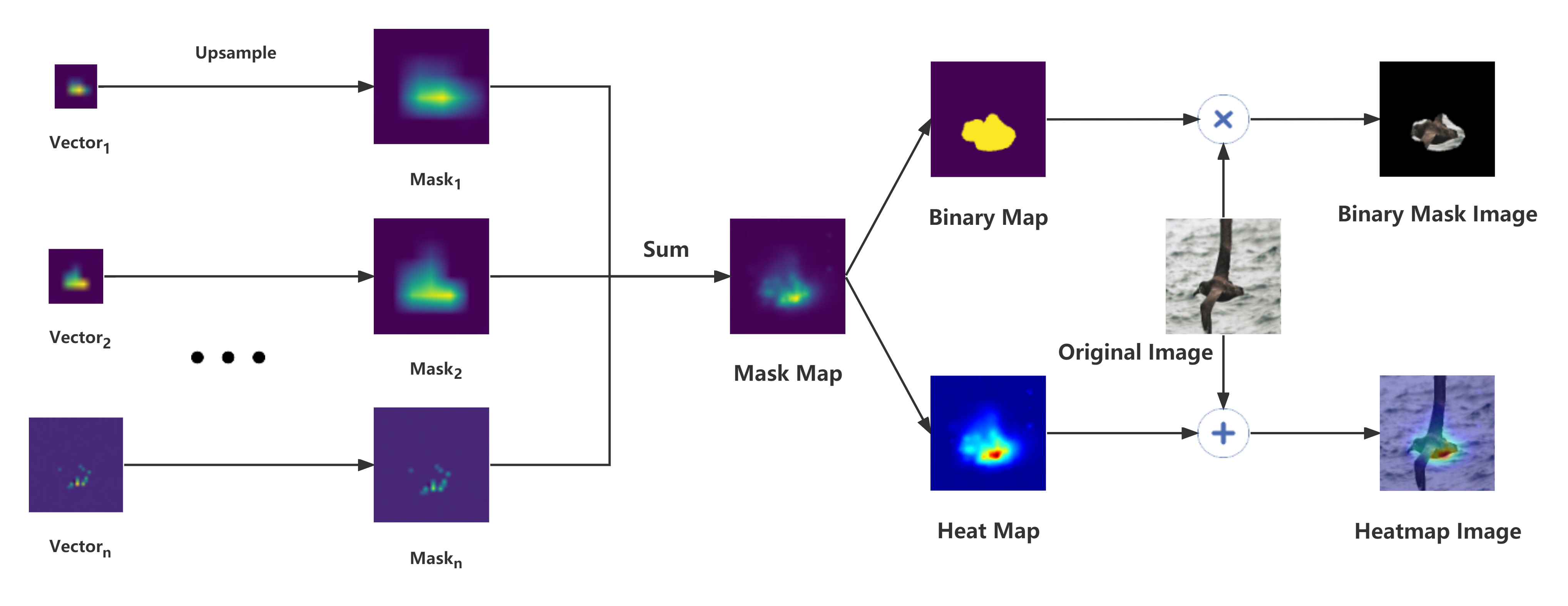} 
	\caption{The pipeline of pre-trained mask feature vectors to generate binary mask image and heatmap image for the neural network classification decisions.}
	\label{fig4}
\end{figure*}

\subsection{Training Process}
Note: The original image input data is ($X$, $Y$), where $X\in R^{H\times W\times C}$ is the input image and $Y$ is the classification label. The neural network for the classification task is denoted as $f$, and $t$ represents the activation position selected by the activation selector. The Mask Generator generates $N$ mask feature vectors $D\{d_{i}\}^{N}_{i=1}$ for each mask $M_{i}=g(d_{i})$. The activation of the original image $X$ and the mask image $M_{i}X$ at the position $t$ after incorporating $f$ are expressed by (3) and (4), respectively.
\begin{equation}
P^{i}_{1,t}=f_{t}(X)
\end{equation}
\begin{equation}
P^{i}_{2,t}=f_{t}(M_{i}X)
\end{equation}
The consistency loss, regularization loss of mask feature vector $d_{i}$ and total loss are expressed by (5), (6) and (7).

\begin{equation}
Loss^{i}_{consistency}=||P^{i}_{1,t}-P^{i}_{2,t}||^2
\end{equation}

\begin{equation}
Loss^{i}_{l_{1}}=\frac{1}{|a_{i}b_{i}|}||d_{i}||_{1}=\frac{1}{|a_{i}b_{i}|}\sum_{u=1}^{a_i}{\sum_{v=1}^{b_i}|d_{iuv}|}
\end{equation}

\begin{equation}
Loss^{i}_{total}=Loss^{i}_{consistency}+\lambda_{i}Loss^{i}_{l_{1}}
\end{equation}

\begin{figure}[t]
	\centering
	\includegraphics[width=0.9\columnwidth]{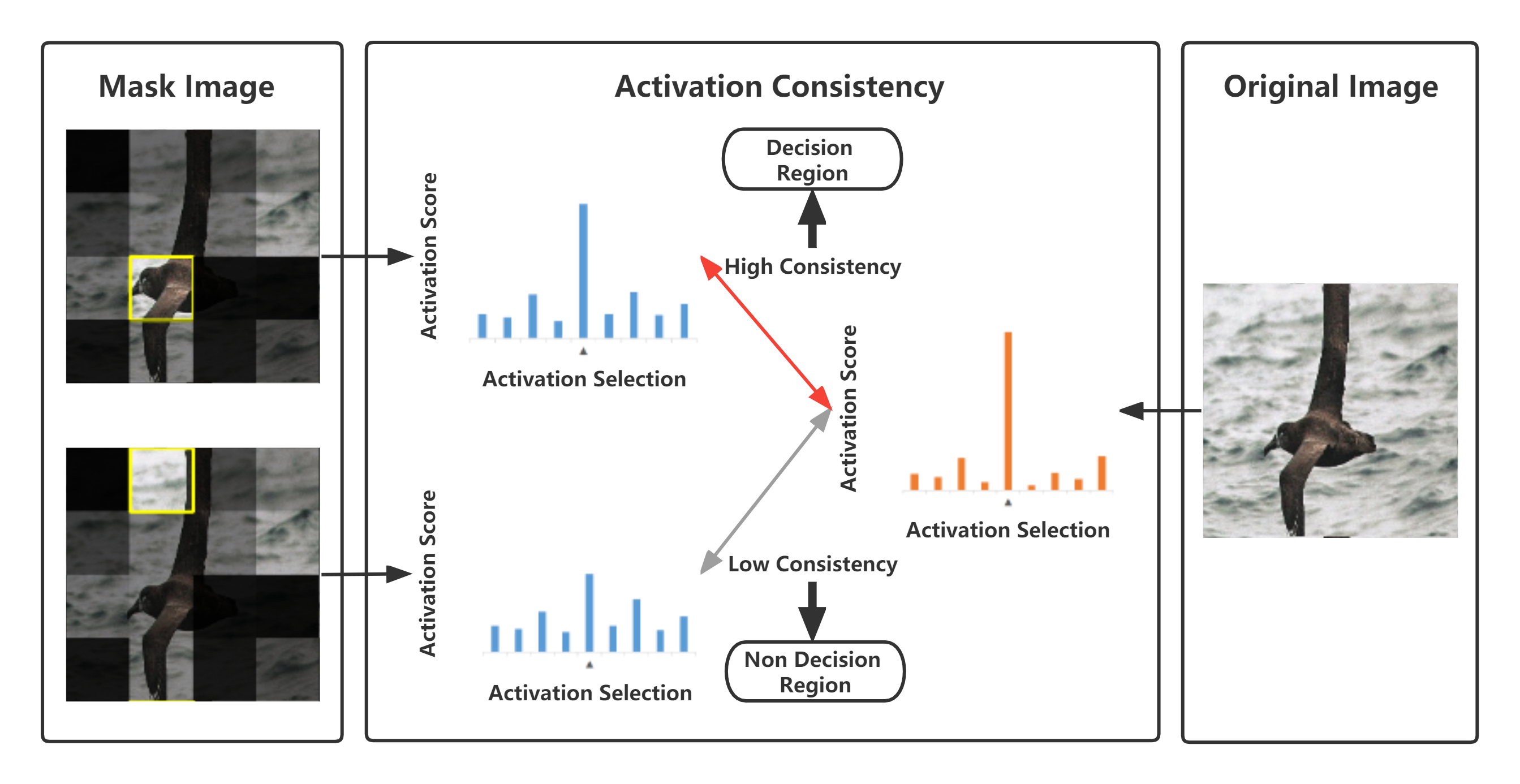} 
	\caption{The mask training process. The decision regions with high activation have high mask values, and irrelevant regions with low activation have low mask values.}
	\label{fig5}
\end{figure}

As show in Figure 5, we train each $d_{i}$ by minimizing (7).

\subsection{Activation Map Generation}
The operation flow is shown in Figure 4. According to trained $D\{d_{i}\}^{N}_{i=1}$ and $M_{i}=g(d_{i})$, we can get $\{M_{i}\}^{N}_{i=1}$ for $X$ and $f$. The binary mask is $M^{b}=\{\sum_{i=1}^{N}M_{i}\geq\gamma\}$, where $\gamma$ is the threshold, $\{\cdot\}$ represents the truth function, 1 if the function is true, otherwise 0.
The activation heatmap mask is $M^{h}=Normalize(\{\sum_{i=1}^{N}M_{i}\geq\gamma\}\sum_{i=1}^{N}M_{i})$. The original image $X$, the $\alpha$, $\beta$ and $\gamma$ are hyperparameters, the activation heatmap image and the binary mask image are $M_{MDM}^{h}$, $M_{MDM}^{b}$, respectively.

Activation heatmap image:
\begin{equation}
M_{MDM}^{h}=\alpha X + \beta M^{h}
\end{equation}

Binary mask image:
\begin{equation}
M_{MDM}^{b}=M^{b}X
\end{equation}

\subsection{Feasibility Proof of Algorithm}
Proposition $P$: By minimizing the objective function (1), the feature mask vector $d_{i}$ can be trained effectively, and then the upsampled $M_{i}$ can mask the classification decision area of image X. The more important decision-making areas are masked lower, and less important areas are more masked.

The following provides a mathematical proof that the above proposition $P$ holds under certain assumptions.
Note: $Z$ represents the region in figure $X$, and $f_{p}(Z)$ represents the activation of the neural network $f$ at $p$ when the data of the region $Z$ is taken as an input.

Let: $I(Z)=kf_{p}(Z)$, where $k$ is a constant greater than zero, $I(Z) \in [0,1]$. $I(Z)$ represents the amount of information contributed by region $Z$ to $p$ activation.

Assumption 1: $z_{1}$ and $z_{2}$ are two regions of the mask feature vector under investigation. When the corresponding regions on the original image do not intersect, it is considered that information $I$ of the contribution of the two regions to activation $f_{p}$ is irrelevant. $g$ is the upsampling function.

$z_{1}$ and $z_{2}$ are the two regions of $d_{i}$, $i \in {1,2,...,N}$.

if $g(z_{1}) \cap g(z_{2})=\varnothing$, then 

\begin{equation}
I(z_{1}+z_{2})=I(z_{1})+I(z_{2})
\end{equation}

Assumption 2: The greater the contribution of the investigation area to the activation, the greater the contribution to the information increment.

$z_{1}$ and $z_{2}$ are the two regions of $d_{i}$, $i \in {1,2,...,N}$, $m\in [0,1]$. if $I(z_{1})<I(z_{2})$, then

\begin{equation}
0\leq \frac{\partial I(mz_{1})}{\partial m} < \frac{\partial I(mz_{2})}{\partial m}
\end{equation}

(1) It can be abbreviated as the following formula:

\begin{equation}
L(m,z) = [f_{p}(z)-f_{p}(mz)]^2 + \lambda m
\end{equation}
$z$ is all areas of $d_{i}$, and $m$ is the corresponding mask value on it.

Proposition $Q$: When $L$ in (12) takes the minimum value, the region mask $m$ with higher activation value in $d_{i}$ is larger, and the region mask $m$ with lower activation value is smaller.

Expressed in mathematical notation: $z_{1}$ and $z_{2}$ represent two disjoint regions of  $d_{i}$, $m_{1}$, $m_{2}$ are mask value on $z_{1}$, $z_{2}$.

$\forall z_{1}, z_{2}$, if $I(z_{1}) < I(z_{2})$, then $m_{1} \leq m_{2}$.

Obviously, the original proposition $P$ is equivalent to the proposition $Q$. The following proves $Q$.

reductio ad absurdum:

If $L$ has obtained the minimum value, and $\exists z_{1}$, $z_{2}$ satisfy: 

$I(z_{1}) < I(z_{2})$ and $m_{1}>m_{2}$.

Define: $z(d_{i})$ means all areas on $d_{i}$, let: $z_{other}=z(d_{i})-z_{1}-z_{2}$, $m_{other}$ represents the mask value of $z_{other}$. Obviously: $g(z_{1}) \cap g(z_{2})=\varnothing$, $g(z_{1}) \cap g(z_{other})=\varnothing$ and $g(z_{2}) \cap g(z_{other})=\varnothing$.

$L(m,z)=L(z_{1}, m_{1}, z_{2}, m_{2}, z_{other}, m_{other})=[f_{p}(z_{1}+z_{2}+z_{other})-f_{p}(m_{1}z_{1}+m_{2}z_{2}+z_{other})]^2+\lambda (m_{1}+m_{2}+m_{other})$.

Let:$L^{'}(m,z)=L(z_{1}, m_{2}, z_{2}, m_{1}, z_{other},m_{other})=[f_{p}(z_{1}+z_{2}+z_{other})-f_{p}(m_{2}z_{1}+m_{1}z_{2}+z_{other})]^2+\lambda (m_{2}+m_{1}+m_{other})$,

$S \triangleq \frac{1}{k}[2f_{p}(z_{1}+z_{2}+z_{other}) - f_{p}(m_{1}z_{1}+m_{2}z_{2}+z_{other}) - f_{p}(m_{2}z_{1}+m_{1}z_{2}+z_{other})] $.

$L^{'}(m,z)-L(m,z)=(kS)[f_{p}(m_{1}z_{1}+m_{2}z_{2}+z_{other})-f_{p}(m_{2}z_{1}+m_{1}z_{2}+z_{other})]$

=$S[I(m_{1}z_{1}) + I(m_{2}z_{2}) + I(z_{other}) - I(m_{2}z_{1}) - I(m_{1}z_{2}) - I(z_{other})]$

=$S \int_{m_{2}}^{m_{1}}[\frac{\partial I(mz_{1})}{\partial m} - \frac{\partial I(mz_{2})}{\partial m}]dm < 0$

$\therefore L^{'}(m,z)<L(m,z)$, contradicting with the minimum value of $L$. Proposition $Q$ is True. Therefore, Proposition $P$ is True.

\begin{figure}[t]
	\centering
	\includegraphics[width=1.0\columnwidth]{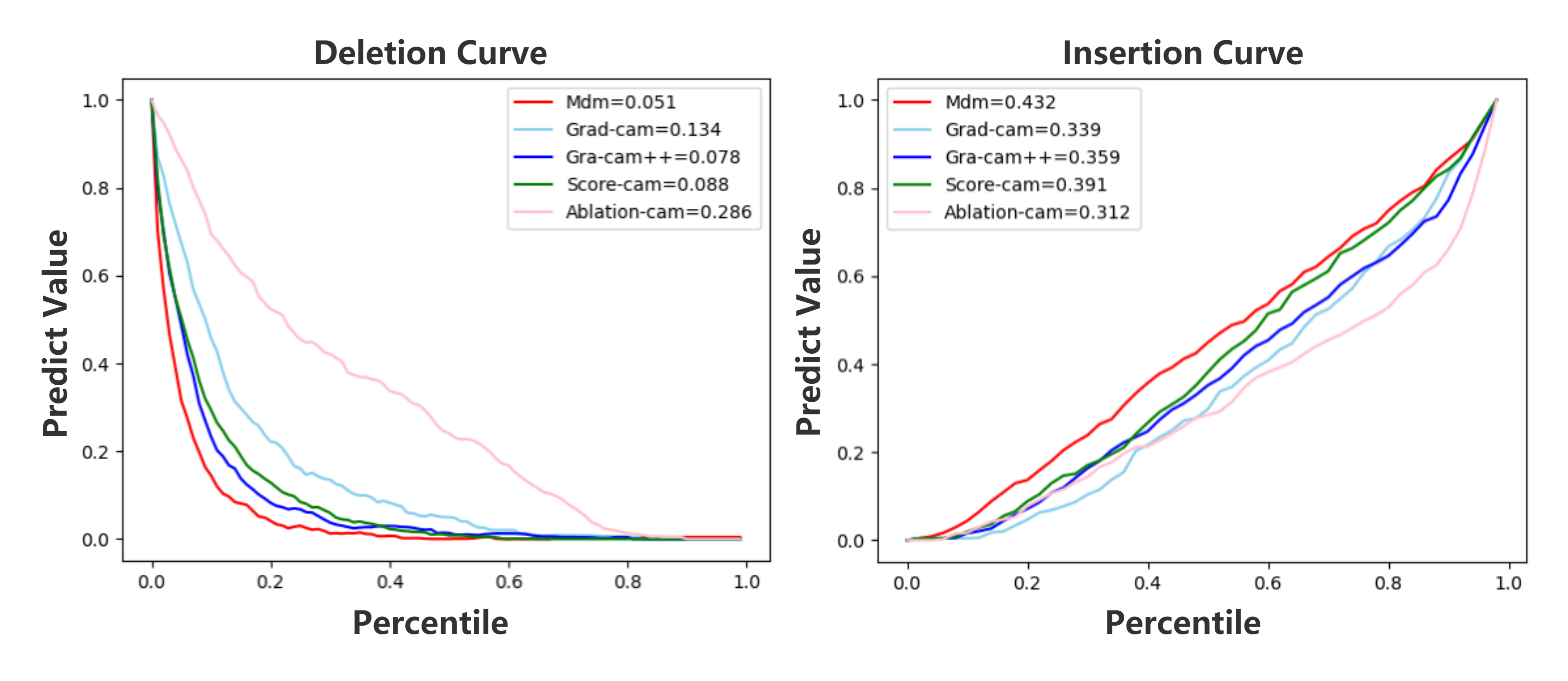} 
	\caption{Deletion and Insertion curves generated under Grad-CAM, Grad-CAM++, Score-CAM, Ablation-CAM and MDM methods in the same neural network.}
	\label{fig6}
\end{figure}

\section{Experiment}

\subsection{Datasets}

\subsubsection{CUB200-2011}It \cite{wah2011caltech} is a bird dataset for testing of fine-grained classification, where each image has a bird of a specific class. There are a total of 11788 bird images, including 200 birds of different categories. We randomly selected 5 images from each of the 200 classes in the test set to form 1000 images to experiment. The MDM method proposed in this paper is compared with some existing methods \cite{selvaraju2017grad,chattopadhay2018grad,wang2020score,ramaswamy2020ablation} on the search effect of classification basis.

\subsubsection{NIH-Chest-X-Ray}It \cite{wang2017chestx} is a publicly available chest X-ray dataset. Among them, there are 880 pictures in the test set with 984 bounding boxes, which frame the corresponding disease positions in the pictures. We use ProtoPNet and XProtoNet trained by the training method presented in \cite{kim2021xprotonet} as the pre-trained neural networks, and compare the prototype search method of the original paper with our proposed MDM in the network. We observed the coincidence of the searched activation area and the marked bounding boxes, and compared the accuracy of the two methods in finding prototypes.

\subsubsection{ImageNet}By setting the neural networks pre-trained on ImageNet \cite{deng2009imagenet} to predict the same images from ImageNet \cite{deng2009imagenet} as different classes, we test whether the MDM can find the valid decision area for classification for various neural networks \cite{simonyan2014very,he2016deep,huang2017densely,dosovitskiy2020image,liu2021swin}. And we visualized the CAM for validation.

\begin{figure}[t]
	\centering
	\includegraphics[width=0.9\columnwidth]{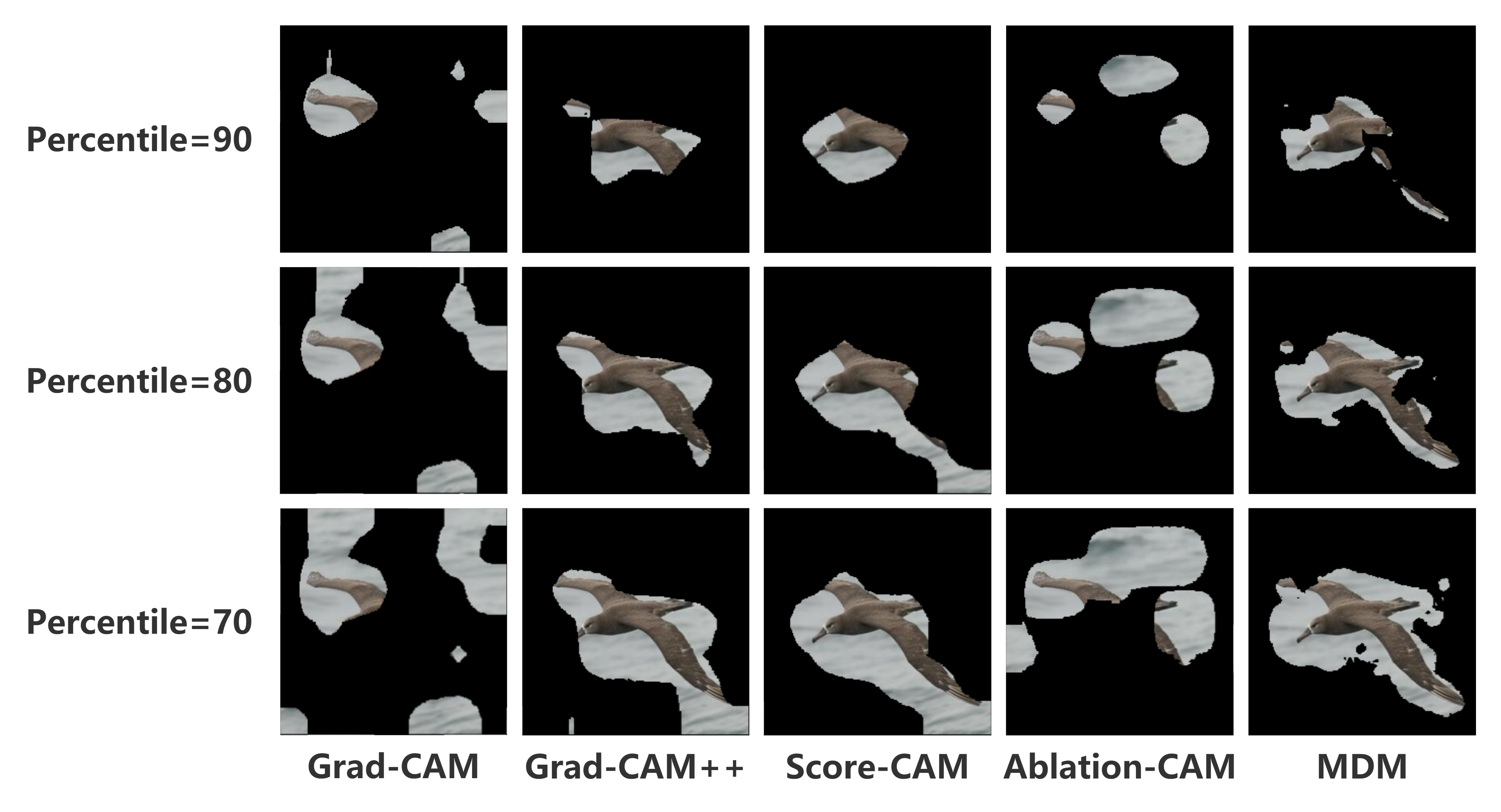} 
	\caption{Visualization of binary mask image of neural network classification decision region. The mask thresholds of the binary mask image under different methods are set to the activation percentages of 90, 80 and 70, respectively.}
	\label{fig7}
\end{figure}

\begin{table}[t]
	\centering
	\begin{tabular}{l|c|c|c|c}
		\hline \diagbox{$Mod$}{$Eva$} & AD(\%) & AI(\%) & Deletion & Insertion\\
		\hline
		grad-cam & 95.82 & 0.43 & 0.134 & 0.339 \\
		\hline
		grad-cam++ & 92.85 & 0.85 & 0.078 & 0.359 \\
		\hline
		score-cam & 91.18 & 1.71 & 0.088 & 0.391 \\
		\hline
		ablation-cam & 96.69 & 0.42 & 0.286 & 0.312 \\
		\hline
		mdm(ours) & \textbf{88.82} & \textbf{2.14} & \textbf{0.051}& \textbf{0.432}\\
		\hline
	\end{tabular}
	\caption{Evaluated results on Recognition, lower is better in Average Drop(AD), and higher is better in Average Increase(AI). Comparative evaluation in terms of Deletion (lower is better) and Insertion (higher is better) scores. Eva and Mod represent Evaluation and Model, respectively.}
	\label{table1}
\end{table}

\begin{table}[t]
	\centering
	\begin{tabular}{|l|c|c|c|c|}
		\hline
		Dataset & \multicolumn{4}{c|}{CUB200-2011} \\
		\hline
		Evaluation & DICE & IOU & PPV & SENS\\
		\hline
		ProtoPNet & 0.432 & 0.287 & 0.645 & 0.359 \\
		\hline
		ProtoPNet\&M(ours) & \textbf{0.516} & \textbf{0.366} & \textbf{0.738} & \textbf{0.442} \\
		\hline
		Dataset & \multicolumn{4}{c|}{NIH chest X-ray} \\
		\hline
		Evaluation & DICE & IOU & PPV & SENS\\
		\hline
		ProtoPNet & 0.256 & 0.158 & 0.263 & 0.428 \\
		\hline
		ProtoPNet\&M(ours) & \textbf{0.283} & \textbf{0.187} & \textbf{0.278} & \textbf{0.486} \\
		\hline	
		Dataset & \multicolumn{4}{c|}{NIH chest X-ray} \\
		\hline
		Evaluation & DICE & IOU & PPV & SENS\\
		\hline
		XProtoNet & 0.120 & 0.068 & 0.070 & \textbf{0.915} \\
		\hline
		XProtoNet\&M(ours) & \textbf{0.125} & \textbf{0.071} & \textbf{0.099} & 0.347 \\	
		\hline
	\end{tabular}
	
	\caption{Comparison of the prototype lookup performance of ProtoPNet and XProtoNet before and after using MDM(M).}
	\label{table2}
\end{table}

\begin{algorithm}[tb]
	\caption{Multiple Dynamic Masks}
	\label{alg:algorithm}
	\textbf{Input}: Image $X_{0}$, Neural Network $f(x)$, Activation Position $t$, Upsample Function $g(x)$, Loss Function $L$.\\
	\textbf{Output}: Heatmap $M^{h}$, Binary Mask $M^{b}$, Heatmap Image $M^{h}_{MDM}$, Binary Mask Image $M^{b}_{MDM}$.\\
	\textbf{Parameter}: Weight $\{\lambda_{i}\}^{N}_{i=1}$, Mask Vectors $\{d_{i}\}^{N}_{i=1}$, Epochs $C$, Learning Rate $\eta$, Threshold $\gamma$, Mix $\alpha$, $\beta$.
	\begin{algorithmic}[1] 
		\STATE Initialize $\{d_{i}\}^{N}_{i=1}$ each element is 0.5
		\STATE $A^{t} \gets f_{t}(X_{0})$
		\STATE \textbf{for} $i=1$ \textbf{to} $N$ \textbf{do}
		\STATE \,\,\,\,\,\,\,\, \textbf{for} $j=1$ \textbf{to} $C$ \textbf{do}
		\STATE \,\,\,\,\,\,\,\,\,\,\,\,\,\,\,\, $M_{i} \gets g(d_{i})$
		\STATE \,\,\,\,\,\,\,\,\,\,\,\,\,\,\,\, $A^{t}_{i} \gets f_{t}(M_{i} \cdot X_{0})$
		\STATE \,\,\,\,\,\,\,\,\,\,\,\,\,\,\,\, $L_{c} \gets L(A^{t}, A^{t}_{i})$
		\STATE \,\,\,\,\,\,\,\,\,\,\,\,\,\,\,\, $L_{d} \gets ||d_{i}||_{1}$
		\STATE \,\,\,\,\,\,\,\,\,\,\,\,\,\,\,\, $L_{t} \gets L_{c} + \lambda_{i}L_{d}$
		\STATE \,\,\,\,\,\,\,\,\,\,\,\,\,\,\,\, $\theta_{d_{i}} \gets \theta_{d_{i}} - \eta \frac{\partial L_{t}}{\partial \theta_{d_{i}}}$
		\STATE \,\,\,\,\,\,\, \textbf{end for}
		\STATE \textbf{end for}
		\STATE Initialize $M^{F}$ to zero mask
		\STATE \textbf{for} $i=1$ \textbf{to} $N$ \textbf{do}
		\STATE \,\,\,\,\,\,\,\, $M^{F} \gets M^{F} + g(d_{i})$
		\STATE \textbf{end for}
		\STATE $M^{b} = \{M^{F} \geq \gamma\}$
		\STATE $M^{h} = M^{b} \cdot M^{F}$
		\STATE Normalize the Heatmap $M^{h}$
		\STATE $M_{MDM}^{h}=\alpha X_{0} + \beta M^{h}$
		\STATE $M^{b}_{MDM} = M^{b} \cdot X_{0}$
		\STATE \textbf{return} $M^{h}$, $M^{b}$, $M^{h}_{MDM}$, $M^{b}_{MDM}$
	\end{algorithmic}
\end{algorithm}

\subsection{Evaluation}
For evaluating the search performance of the decision activation area, we choose the following indicators: Average Drop and Average Increase proposed by \cite{chattopadhay2018grad}; and Deletion and Insertion proposed by \cite{petsiuk2018rise}. The above four evaluation indicators are widely used for comparing the performance of CAM. Dice Coefficient, IOU, PPV, Sensitivity of the activation area and the segmented foreground or the detection area. The above eight evaluation indicators are compared. The Average Drop is expressed as: $\sum_{i=1}^{N} \frac{\max (0,Y^{c}_{i}-O^{c}_{i})}{Y^{c}_{i}} \times 100$. The Average Increase is expressed as: $\frac{1}{N}\sum_{i=1}^{N}Sign(Y^{c}_{i}<O^{c}_{i})$. $Y^{c}_{i}$ represents the predicted score of class $c$ in the original image $i$, and $O^{c}_{i}$ represents predicted score of the class $c$ with explained map obtained after the original image is masked. $Sign$ represents an indicator function that returns 1 if the input is true. We removed certain percentile pixels of the original image to generate a explained map.

The Deletion and Insertion metrics are based on the CAM to remove and insert pixels from the original image in descending order of activation value, respectively, and generate the area under the probability curve(AUC) depicted by the predicted probability result of the picture after removal or insertion. Lower deletion score is better and higher insertion score is better.

We believe that the neural network should have large activations for areas that are effective for classification predictions, and low activations for areas that are ineffective for classification predictions. We set the activation value of a certain percentile of CAM as threshold, and we use the threshold to generate the binary mask. We adopt the Dice Coefficient, IOU, PPV and Sensitivity calculated by the foreground image and the binary mask or the real bounding box and the binary mask as evaluation metrics.

\subsection{Experimental Details}
All mask vectors of MDM are initialized to 0.5 per element. The size of cropped images for the CUB dataset is $224 \times 224$ and that for the NIH dataset is  $512 \times 512$. We set $N=27$, $D\{d_{i}\}^{27}_{i=1}$, $d_{i} \in R^{a_{i} \times b_{i} \times 1}$, $a_{i}=b_{i}=5+i, 1 \leq i \leq 27$. $g(\cdot)$ adopt bilinear upsampling and normalization. Set the threshold $\gamma = 5$, $\alpha = 0.5$, $\beta = 0.3$. In the CUB dataset, set $\lambda_{i}=1e2$, and in the NIH dataset, set $\lambda_{i}=1e3$, where $i \in \{1,2,...,N\}$. We use the Adam optimizer, the learning rate $lr=3e-3$. Set activation area of ProtoPNet and XProtoNet separately to the pixels in the top 10\% and top 20\% of the activation values. Each mask is trained for 2000 iterations. The network is trained on the corresponding dataset after it was pre-trained on ImageNet \cite{deng2009imagenet}. In the CUB200-2011 and the NIH-Chest X-ray dataset, the number of prototypes is set to 10 and 3, respectively. Arbitrarily select a prototype as CAM. Set the prototype bounding box to include the pixels in the top 5\% of the activation values. All models are trained on 1 2080Ti GPU.

\subsection{Comparison with Baselines}
In the CUB200-2011 and NIH dataset, ResNet and DenseNet are used as the pre-trained networks. Grad-CAM \cite{selvaraju2017grad}, Grad-CAM++ \cite{chattopadhay2018grad}, Score-CAM \cite{wang2020score} and Ablation-CAM \cite{ramaswamy2020ablation} are some current the most advanced activation map methods which are tested. The MDM method is compared with them on the above evaluation indicators. We test and compare the prototype lookup performance of ProtoPNet and XProtoNet before and after using the MDM method. When comparing the above methods, the parameters of selected network are fixed.

\subsection{Visualization}
We mixed the activation heatmap generated by MDM with the original image as a visual result. We tested on different neural networks pre-trained on ImageNet: ResNet50, VGG19, DenseNet121, VIT-base (VIT-B) and Swin-Transformer-base (Swin-B).

\begin{figure}[t]
	\centering
	\includegraphics[width=0.9\columnwidth]{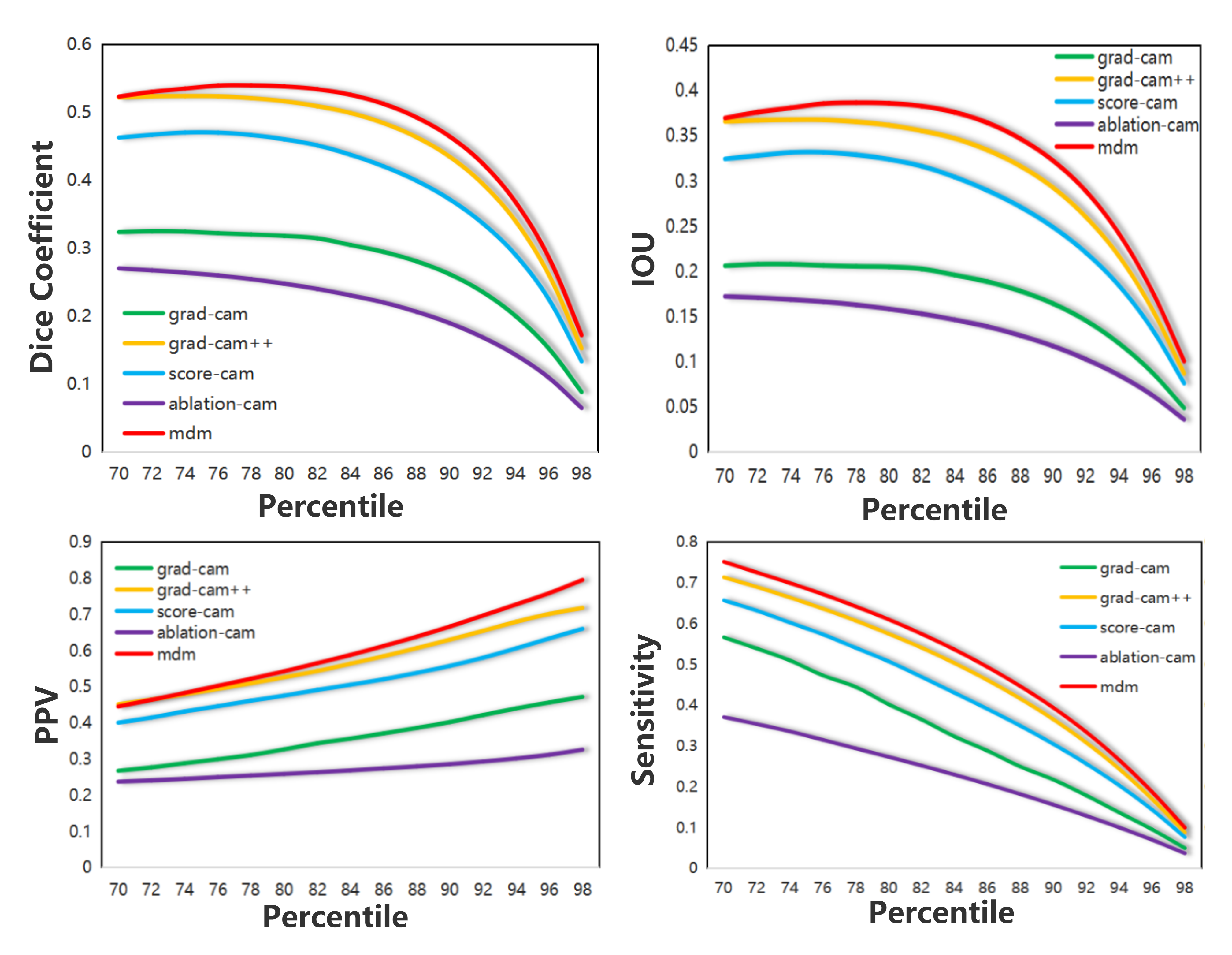} 
	\caption{In convolutional neural networks, five methods corresponding to the curves of Dice Coefficient, IOU, PPV and Sensitivity when the importance percentile of the masked image pixels are masked from 70 to 99.}
	\label{fig8}
\end{figure}

\begin{figure}[t]
	\centering
	\includegraphics[width=0.9\columnwidth]{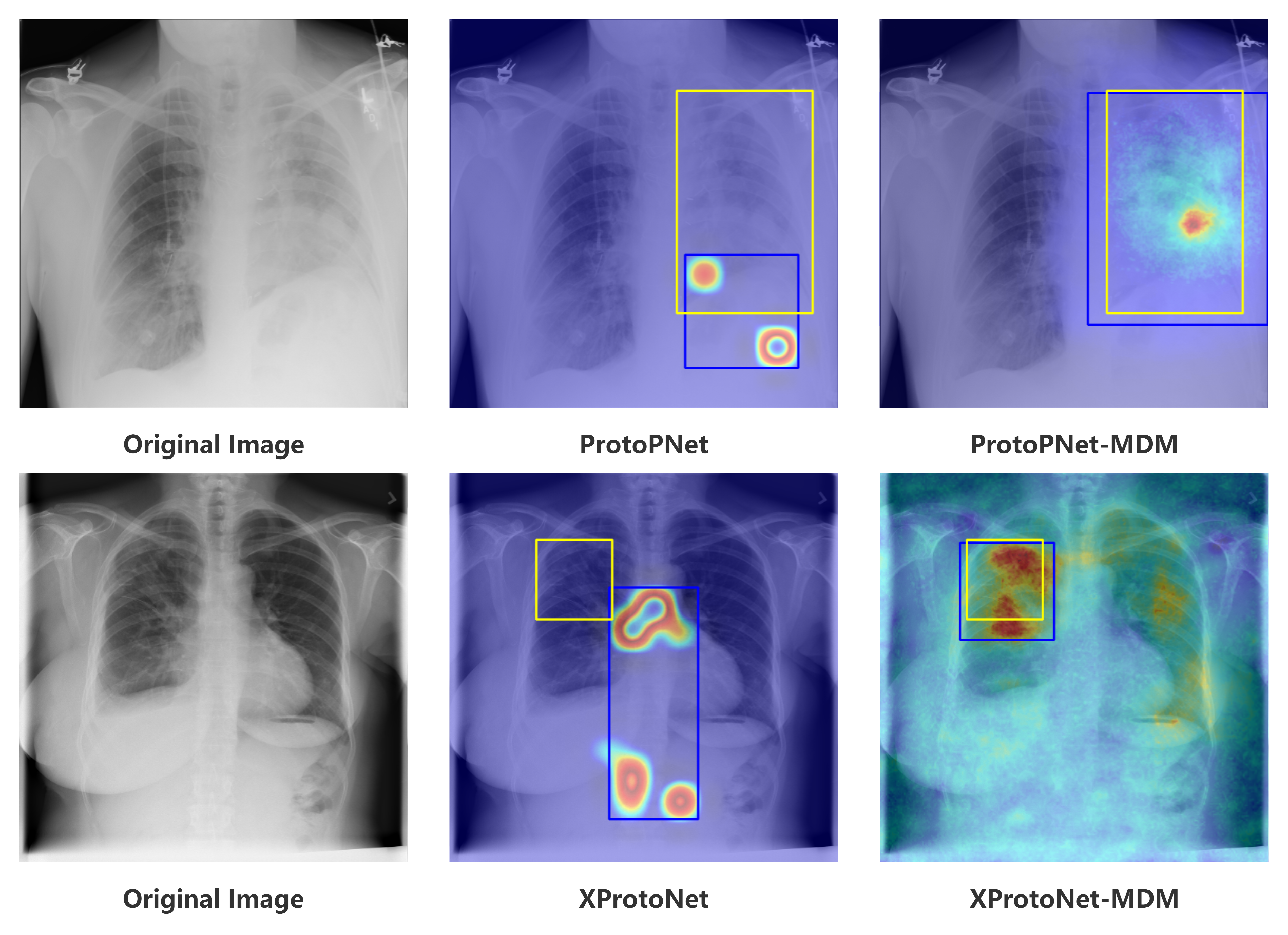} 
	\caption{Visualization of the results of ProtoPNet and XProtoNet before and after using MDM. The yellow box represents the real lesion area, and the blue box represents the  bounding box of lesion area found by above methods.}
	\label{fig9}
\end{figure}

\begin{figure}[t]
	\centering
	\includegraphics[width=0.9\columnwidth]{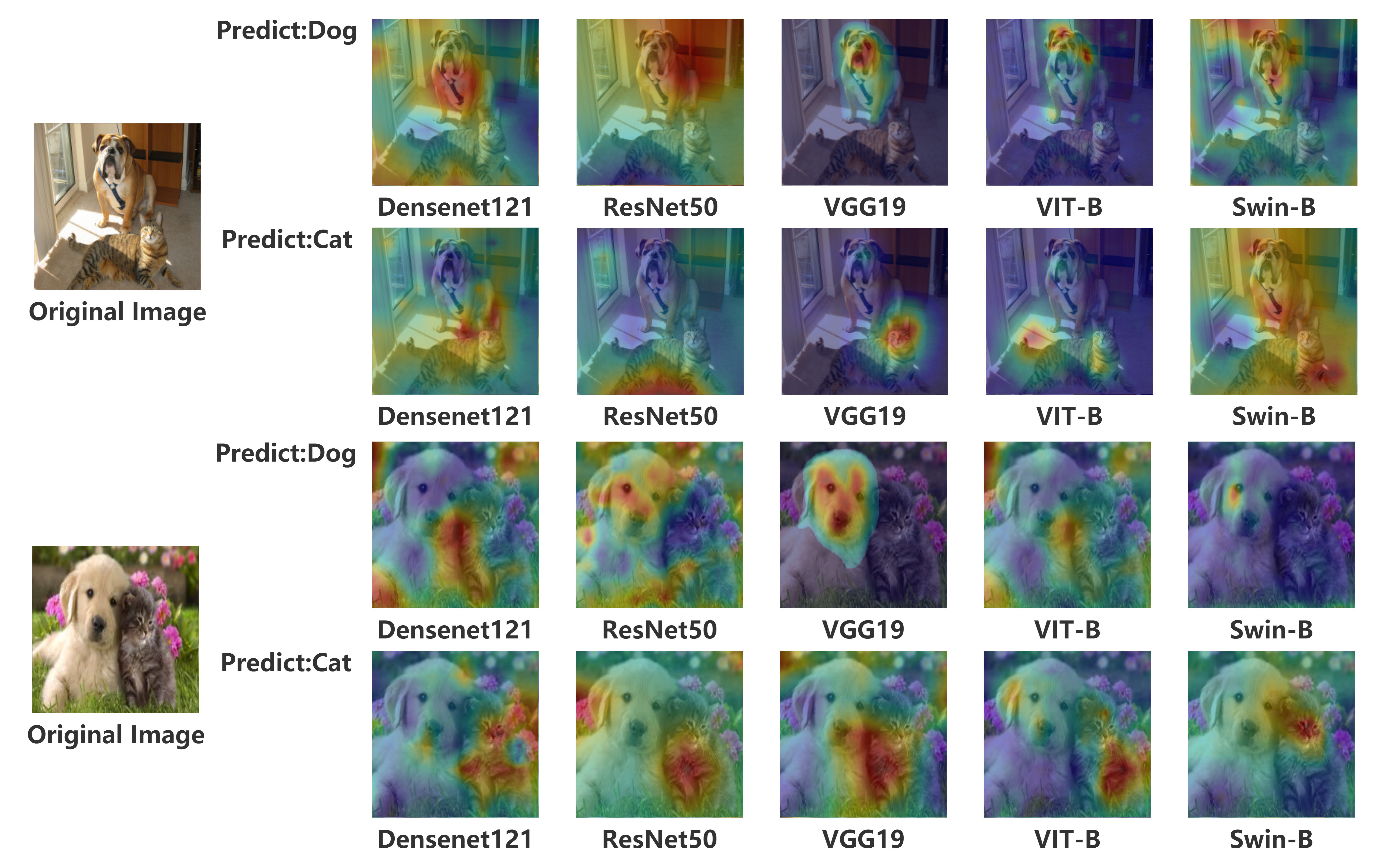} 
	\caption{MDM generates a visualization of the areas of interest when making various neural network decisions.}
	\label{fig10}
\end{figure}

\section{Discussion}
Notice in Figure 7 and Figure 8. On the same deep learning model, comparing with the most advanced CAM methods, specifically Grad-CAM, Grad-CAM++, Score-CAM and Ablation-CAM. When the occlusion pixel index increases from 70\% to 99\%, the MDM achieves the best to find foreground area for classification among the reserved pixels. MDM achieves the state-of-the-art on foreground lookup.

Table 1 and Figure 6 show the CAM performance evaluation indicators used widely. MDM achieves the state of the art in the above indicators. The pixel point area concerned by MDM is the most favorable point for the neural network to classify in the pixel point area found by other methods.

Notice in Table 2, in the bird dataset after using the MDM to find the prototype, ProtoPNet, the evaluation indicators of Dice Coefficient, IOU, PPV, Sensitivity are increased by 19.3\%, 27.1\%, 10.4\% and 22.9\%, respectively. MDM can greatly improve the search performance of interpretable neural network classification decision basis prototypes.

According to Figure 9 and Table 2, when MDM is adopted, ProtoPNet improves the four evaluation indicators of Dice Coefficient, IOU, PPV, and Sensitivity for the prototype search of pathological regions that explain the classification basis. The improvements were 10.2\%, 18.3\%, 5.8\% and 13.5\%. XProtoNet also has a certain increase in most indicators. MDM can greatly improve the performance of the interpretable neural network in finding the lesion area.

In Figure 10, the visualization experiment shows that MDM can well find the decision when any model makes a classification. The MDM method has the generality, and it can be easily used in neural networks of any structure.

\section{Conclusion}
Our proposed Multiple Dynamic Masks (MDM) can point out important activation regions for neural network classification, and it represents an interpretable basis for neural network classification decisions. The reasoning process of MDM conforms to human cognition and it is interpretable. The MDM method is based on learning, it can adaptively find important activation regions for classification, and the search performance of MDM in neural network classification decision regions achieves the state-of-the-art. It can be well used in interpretable neural networks and improved the prototype search performance. MDM is universal, which can be applied to most of advanced neural networks.

\bibliography{mdm_pyt}

\begin{thebibliography}{10}

\bibitem{chattopadhay2018grad}
Aditya Chattopadhay, Anirban Sarkar, Prantik Howlader, and Vineeth~N
  Balasubramanian.
\newblock Grad-cam++: Generalized gradient-based visual explanations for deep
  convolutional networks.
\newblock In {\em 2018 IEEE winter conference on applications of computer
  vision (WACV)}, pages 839--847. IEEE, 2018.

\bibitem{chen2019looks}
Chaofan Chen, Oscar Li, Daniel Tao, Alina Barnett, Cynthia Rudin, and
  Jonathan~K Su.
\newblock This looks like that: deep learning for interpretable image
  recognition.
\newblock {\em Advances in neural information processing systems}, 32, 2019.

\bibitem{deng2009imagenet}
Jia Deng, Wei Dong, Richard Socher, Li-Jia Li, Kai Li, and Li~Fei-Fei.
\newblock Imagenet: A large-scale hierarchical image database.
\newblock In {\em 2009 IEEE conference on computer vision and pattern
  recognition}, pages 248--255. Ieee, 2009.

\bibitem{dosovitskiy2020image}
Alexey Dosovitskiy, Lucas Beyer, Alexander Kolesnikov, Dirk Weissenborn,
  Xiaohua Zhai, Thomas Unterthiner, Mostafa Dehghani, Matthias Minderer, Georg
  Heigold, Sylvain Gelly, et~al.
\newblock An image is worth 16x16 words: Transformers for image recognition at
  scale.
\newblock {\em arXiv preprint arXiv:2010.11929}, 2020.

\bibitem{hashimoto2020multi}
Noriaki Hashimoto, Daisuke Fukushima, Ryoichi Koga, Yusuke Takagi, Kaho Ko, Kei
  Kohno, Masato Nakaguro, Shigeo Nakamura, Hidekata Hontani, and Ichiro
  Takeuchi.
\newblock Multi-scale domain-adversarial multiple-instance cnn for cancer
  subtype classification with unannotated histopathological images.
\newblock In {\em Proceedings of the IEEE/CVF conference on computer vision and
  pattern recognition}, pages 3852--3861, 2020.

\bibitem{he2016deep}
Kaiming He, Xiangyu Zhang, Shaoqing Ren, and Jian Sun.
\newblock Deep residual learning for image recognition.
\newblock In {\em Proceedings of the IEEE conference on computer vision and
  pattern recognition}, pages 770--778, 2016.

\bibitem{huang2017densely}
Gao Huang, Zhuang Liu, Laurens Van Der~Maaten, and Kilian~Q Weinberger.
\newblock Densely connected convolutional networks.
\newblock In {\em Proceedings of the IEEE conference on computer vision and
  pattern recognition}, pages 4700--4708, 2017.

\bibitem{kim2021xprotonet}
Eunji Kim, Siwon Kim, Minji Seo, and Sungroh Yoon.
\newblock Xprotonet: diagnosis in chest radiography with global and local
  explanations.
\newblock In {\em Proceedings of the IEEE/CVF conference on computer vision and
  pattern recognition}, pages 15719--15728, 2021.

\bibitem{lin2013network}
Min Lin, Qiang Chen, and Shuicheng Yan.
\newblock Network in network.
\newblock {\em arXiv preprint arXiv:1312.4400}, 2013.

\bibitem{liu2021swin}
Ze~Liu, Yutong Lin, Yue Cao, Han Hu, Yixuan Wei, Zheng Zhang, Stephen Lin, and
  Baining Guo.
\newblock Swin transformer: Hierarchical vision transformer using shifted
  windows.
\newblock In {\em Proceedings of the IEEE/CVF International Conference on
  Computer Vision}, pages 10012--10022, 2021.

\bibitem{liu2022convnet}
Zhuang Liu, Hanzi Mao, Chao-Yuan Wu, Christoph Feichtenhofer, Trevor Darrell,
  and Saining Xie.
\newblock A convnet for the 2020s.
\newblock In {\em Proceedings of the IEEE/CVF Conference on Computer Vision and
  Pattern Recognition}, pages 11976--11986, 2022.

\bibitem{maksoud2020sos}
Sam Maksoud, Kun Zhao, Peter Hobson, Anthony Jennings, and Brian~C Lovell.
\newblock Sos: Selective objective switch for rapid immunofluorescence whole
  slide image classification.
\newblock In {\em Proceedings of the IEEE/CVF Conference on Computer Vision and
  Pattern Recognition}, pages 3862--3871, 2020.

\bibitem{patricio2022explainable}
Cristiano Patr{\'\i}cio, Jo{\~a}o~C Neves, and Lu{\'\i}s~F Teixeira.
\newblock Explainable deep learning methods in medical diagnosis: A survey.
\newblock {\em arXiv preprint arXiv:2205.04766}, 2022.

\bibitem{petsiuk2018rise}
Vitali Petsiuk, Abir Das, and Kate Saenko.
\newblock Rise: Randomized input sampling for explanation of black-box models.
\newblock {\em arXiv preprint arXiv:1806.07421}, 2018.

\bibitem{ramaswamy2020ablation}
Harish~Guruprasad Ramaswamy et~al.
\newblock Ablation-cam: Visual explanations for deep convolutional network via
  gradient-free localization.
\newblock In {\em Proceedings of the IEEE/CVF Winter Conference on Applications
  of Computer Vision}, pages 983--991, 2020.

\bibitem{selvaraju2017grad}
Ramprasaath~R Selvaraju, Michael Cogswell, Abhishek Das, Ramakrishna Vedantam,
  Devi Parikh, and Dhruv Batra.
\newblock Grad-cam: Visual explanations from deep networks via gradient-based
  localization.
\newblock In {\em Proceedings of the IEEE international conference on computer
  vision}, pages 618--626, 2017.

\bibitem{shrikumar2017learning}
Avanti Shrikumar, Peyton Greenside, and Anshul Kundaje.
\newblock Learning important features through propagating activation
  differences.
\newblock In {\em International conference on machine learning}, pages
  3145--3153. PMLR, 2017.

\bibitem{simonyan2014very}
Karen Simonyan and Andrew Zisserman.
\newblock Very deep convolutional networks for large-scale image recognition.
\newblock {\em arXiv preprint arXiv:1409.1556}, 2014.

\bibitem{singh2021interpretable}
Gurmail Singh and Kin-Choong Yow.
\newblock An interpretable deep learning model for covid-19 detection with
  chest x-ray images.
\newblock {\em Ieee Access}, 9:85198--85208, 2021.

\bibitem{singh2021these}
Gurmail Singh and Kin-Choong Yow.
\newblock These do not look like those: An interpretable deep learning model
  for image recognition.
\newblock {\em IEEE Access}, 9:41482--41493, 2021.

\bibitem{sundararajan2017axiomatic}
Mukund Sundararajan, Ankur Taly, and Qiqi Yan.
\newblock Axiomatic attribution for deep networks.
\newblock In {\em International conference on machine learning}, pages
  3319--3328. PMLR, 2017.

\bibitem{szegedy2015going}
Christian Szegedy, Wei Liu, Yangqing Jia, Pierre Sermanet, Scott Reed, Dragomir
  Anguelov, Dumitru Erhan, Vincent Vanhoucke, and Andrew Rabinovich.
\newblock Going deeper with convolutions.
\newblock In {\em Proceedings of the IEEE conference on computer vision and
  pattern recognition}, pages 1--9, 2015.

\bibitem{wah2011caltech}
Catherine Wah, Steve Branson, Peter Welinder, Pietro Perona, and Serge
  Belongie.
\newblock The caltech-ucsd birds-200-2011 dataset.
\newblock 2011.

\bibitem{wang2020score}
Haofan Wang, Zifan Wang, Mengnan Du, Fan Yang, Zijian Zhang, Sirui Ding, Piotr
  Mardziel, and Xia Hu.
\newblock Score-cam: Score-weighted visual explanations for convolutional
  neural networks.
\newblock In {\em Proceedings of the IEEE/CVF conference on computer vision and
  pattern recognition workshops}, pages 24--25, 2020.

\bibitem{wang2017chestx}
Xiaosong Wang, Yifan Peng, Le~Lu, Zhiyong Lu, Mohammadhadi Bagheri, and
  Ronald~M Summers.
\newblock Chestx-ray8: Hospital-scale chest x-ray database and benchmarks on
  weakly-supervised classification and localization of common thorax diseases.
\newblock In {\em Proceedings of the IEEE conference on computer vision and
  pattern recognition}, pages 2097--2106, 2017.

\bibitem{zhou2016learning}
Bolei Zhou, Aditya Khosla, Agata Lapedriza, Aude Oliva, and Antonio Torralba.
\newblock Learning deep features for discriminative localization.
\newblock In {\em Proceedings of the IEEE conference on computer vision and
  pattern recognition}, pages 2921--2929, 2016.

\end{thebibliography}

\end{document}